\documentclass{article}

\date{} 

\usepackage{enumitem}

\usepackage{url}

\usepackage{hyperref} 

\usepackage{spconf,amsmath,graphicx}
\usepackage{amsmath}
\usepackage{mathrsfs}
\usepackage{caption}
\usepackage{amssymb}
\usepackage{mathabx}
\usepackage{algorithm,algorithmic}
\usepackage{multirow}
\usepackage{array}
\usepackage{graphicx}
\usepackage{subfigure}
\usepackage{color}
\usepackage{lipsum}
\usepackage{cite}

\usepackage{epstopdf}

\let\OLDthebibliography\thebibliography
\renewcommand\thebibliography[1]{
  \OLDthebibliography{#1}
  \setlength{\parskip}{0pt}
  \setlength{\itemsep}{3pt plus 0.3ex}
}

\title{Saliency detection for seismic applications using multi-dimensional spectral projections and directional comparisons}
\name{Muhammad Amir Shafiq$^{*\dagger}$, Zhiling Long$^\dagger$, Tariq Alshawi, and Ghassan AlRegib$^\dagger$\thanks{This work is supported by the Center for Energy and Geo Processing (CeGP) at Georgia Institute of Technology and King Fahd University of Petroleum and Minerals.}}
\address{ $^\dagger$Center for Energy and Geo Processing (CeGP) at Georgia Tech and KFUPM\\
School of Electrical and Computer Engineering\\
Georgia Institute of Technology, Atlanta, Georgia, 30332-0250\\
\{amirshafiq$^*$,~zhiling.long,~talshawi,~alregib\}@gatech.edu}
\begin{document}

\onecolumn 

\begin{description}[labelindent=1cm,leftmargin=4cm,style=multiline]

\item[\textbf{Citation}]{M. Shafiq, Z. Long, T. Alshawi, and G. AlRegib, ``Saliency detection for seismic applications using multi-dimensional spectral projections and directional comparisons,'' Proceedings of IEEE International Conference on Image Processing (ICIP), Beijing, China, Sep. 2017.}
\\
\item[\textbf{DOI}]{\url{https://doi.org/10.1109/ICIP.2017.8296322}}
\\
\item[\textbf{Review}]{Date of publication: 22 February 2018}
\\
\item[\textbf{Data and Codes}]{\url{https://ghassanalregibdotcom.files.wordpress.com/2016/10/amir_icip2017_code.zip}}
\\
\item[\textbf{Bib}] {@INPROCEEDINGS\{8296322, \\ 
author=\{M. A. Shafiq and Z. Long and T. Alshawi and G. AlRegib\}, \\ 
booktitle=\{2017 IEEE International Conference on Image Processing (ICIP)\}, \\ 
title=\{Saliency detection for seismic applications using multi-dimensional spectral projections and directional comparisons\}, \\ 
year=\{2017\}, \\ 
pages=\{455-459\}, \\ 
ISSN=\{2381-8549\}, \\ 
month=\{Sep.\}\}
} 
\\

\item[\textbf{Copyright}]{\textcopyright 2017 IEEE. Personal use of this material is permitted. Permission from IEEE must be obtained for all other uses, in any current or future media, including reprinting/republishing this material for advertising or promotional purposes, creating new collective works, for resale or redistribution to servers or lists, or reuse of any copyrighted component of this work in other works.}
\\
\item[\textbf{Contact}]{\href{mailto:zhiling.long@gatech.edu}{zhiling.long@gatech.edu}  OR \href{mailto:alregib@gatech.edu}{alregib@gatech.edu}\\ \url{https://ghassanalregib.com/} \\ }
\end{description}

\thispagestyle{empty}
\newpage
\clearpage
\setcounter{page}{1}

\twocolumn

%
\maketitle
\begin{abstract}
In this paper, we propose a novel approach for saliency detection for seismic applications using 3D-FFT local spectra and multi-dimensional plane projections. We develop a projection scheme by dividing a 3D-FFT local spectrum of a data volume into three distinct components, each depicting changes along a different dimension of the data. The saliency detection results obtained using each projected component are then combined to yield a saliency map. To accommodate the directional nature of seismic data, in this work, we modify the center-surround model, proven to be biologically plausible for visual attention, to incorporate directional comparisons around each voxel in a 3D volume. Experimental results on real seismic dataset from the F3 block in Netherlands offshore in the North Sea prove that the proposed algorithm is effective, efficient, and scalable. Furthermore, a subjective comparison of the results shows that it outperforms the state-of-the-art methods for saliency detection.
\end{abstract}
\vspace{-.2cm}
\section{Introduction}
\label{sec:intro}
\vspace{-.2cm}
Saliency detection aims to highlight the salient regions in images and videos by taking into consideration the biological structure of the human visual system (HVS)~\cite{Borji2013}. Bottom-up saliency detection in videos exploits both spatial and temporal cues to identify visually important features in the data. Features like color, contrast, intensity, flicker, and motion all have been identified as prominent attributes that help HVS to focus processing resources on important elements in the surrounding environment. In contrast, top-down saliency detection known as task-specific visual saliency embeds a priori knowledge such as shape, orientation, size, or one or more templates of desired features into saliency detection framework. These features provide a user-assisted framework, which in turn makes the saliency detection tuneable to desired features from images or videos.

Majority of visual saliency models aim to predict the areas in images or videos that attraction human attention instantly \cite{itti1998model, fang2014video, qin2014integration, li2014visual, koehler2014saliency, ren2015exploiting}. Itti and Baldi \cite{itti2009bayesian} proposed to detect saliency in images by modeling surprise elicited by an observer by measuring the difference between posterior and prior beliefs. Kadir and Brady \cite{kadir2001saliency} proposed to highlight salient regions by modeling scale selection and content description from images. Li \emph{et al.} \cite{li2011saliency} proposed a novel approach for image segmentation based on sparse saliency model and graph cuts. Furthermore, authors in  \cite{borji2015salient,borji2013quantitative,borji2013state} present a comparison of several state-of-the-art models over seven challenging datasets to establish the benchmark for saliency detection. More recently, a saliency detection algorithm, which uses 3D FFT of a non-overlapping window in the spatial and temporal domains of video sequence to compute the spectral energy of the window and compare it with its surrounding regions to construct a saliency map is proposed in~\cite{Long2015}. This method is effective in capturing both temporal and spatial saliency cues in a very fast and compact way. Therefore, based on the method presented in \cite{Long2015}, we propose a new approach for detecting salient objects within seismic volumes.

\begin{figure*}[ht]
  \centering
  \includegraphics[width=16.0cm]{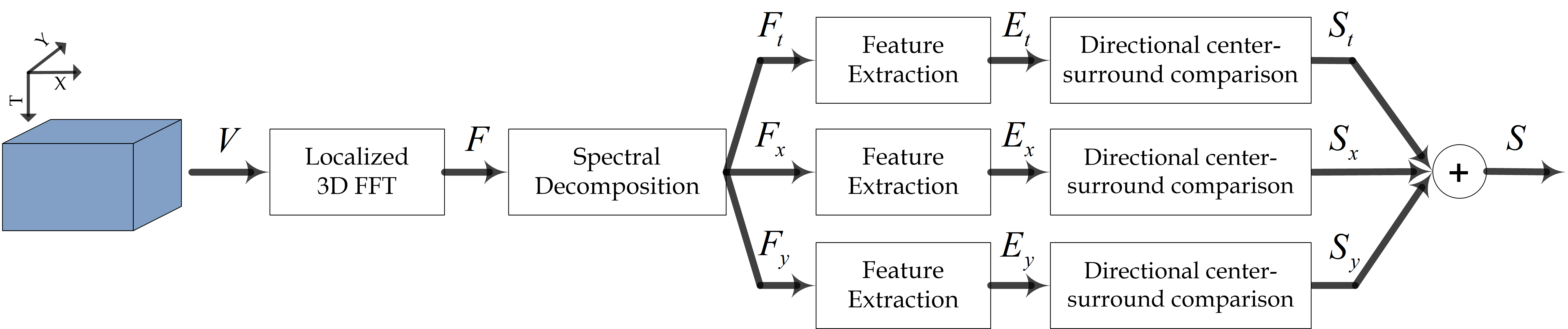}
  \caption{The block diagram of the proposed method.}\label{fig:Block_diag_Proposed}
\end{figure*}

In seismic interpretation, visual saliency is important to predict the human interpreters attention and highlight the areas of interest in seismic sections. To extract useful information from a huge volume of seismic data, interpreters manually delineate important structures, which contain hints about petroleum and gas reservoirs such as salt domes, faults, channels, fractures, pinchouts, anticlines, synclines, and horizons. There are very limited tools available for automatic detection and manual interpretation is becoming extremely time consuming and labor intensive. Therefore, it is important to highlight, in the first stage of interpretation, useful features in seismic images that assist interpreters by directing their attention to the areas, which contain geologically important structures for the entrapment of oil and gas reservoirs. Using visual saliency, we can accomplish this task and make interpreters job relatively easy. Sivarajah \emph{et al.} \cite{sivarajah2014visual} studied various saliency detection algorithms and observed which one closely mimics the interpreter’s visual attention when interpreting the gravity and magnetic data for exploration applications. The study concludes that saliency maps can be used to develop new techniques to compensate or augment biases and guide the interpreter's attention to important areas in images. Similar study to develop a heuristic knowledge of the experts doing interpretation of seismic images is presented in \cite{ahujaexpert}. The authors of \cite{Drissi_Sal, faraklioti2004horizon} proposed algorithms for automated horizons picking by detecting salient features followed by computing pixels entropy and fragments connectivity, respectively. On the other hand, the authors in \cite{Shafiq_ICASSP2016} and \cite{Shafiq_GP2016} proposed novel algorithms for the detection and delineation of salt domes based on visual saliency.

Majority of existing saliency detection algorithms rely on time-space domain for saliency detection. Few schemes based on transform domain have also been proposed in the past for saliency detection, yet it remains as one of the rarely explored domain for saliency detection. For saliency detection in videos, motion-related changes usually occur in the time-domain. However, in seismic data, the variation of facies, faults, salt domes, and different geological features can be observed along all three directions of a 3D seismic data. The utilization of transform domain techniques such as 3D-FFT can capture changes along all three directions in a 3D spectrum. Furthermore, using a top-down approach in saliency calculation, we can enhance the hardly conspicuous features within images and videos. In case of seismic data, certain features such as faults, horizon and sigmoid can be highlighted by defining a template or choosing the size and orientation of saliency calculation in a modified center-surround comparison. Therefore, in this paper, we propose a novel approach for saliency detection, which decomposes a 3D-FFT spectrum into three different components depicting variations along each plane of a 3D volume. Based on obtained spectral decompositions, we apply a modified center-surround model followed by a weighted combination to yield a saliency map of 3D data. Using proposed scheme, we can process visual stimuli in real-time and perform complex processing procedures faster and more efficiently. To show effectiveness of the proposed scheme, we present experimental results on a real dataset from the North Sea, F3 block in the Netherlands and show how proposed algorithm can play an effective role in a computational seismic interpretation process.

\vspace{-.2cm}
\section{Saliency Detection}
\vspace{-.2cm}
In this paper, we develop a novel scheme for saliency detection, which decomposes a 3D-FFT spectrum of data in conjunction with directional center-surround (DCS) model and top-down approach to depict variations in motion along all three dimensions of a 3D data. Given a 3D seismic data volume $\boldsymbol V$ of size $T \times X \times Y$, where $T$ represents time or depth, $X$ represents crosslines, and $Y$ represents inlines, we compute saliency using the block diagram shown in Fig.~\ref{fig:Block_diag_Proposed}.

\begin{figure}[tbp]
\begin{minipage}[b]{.49\linewidth}
  \centering
  \hspace{-1cm}
  \centerline{\includegraphics[width=3.0cm]{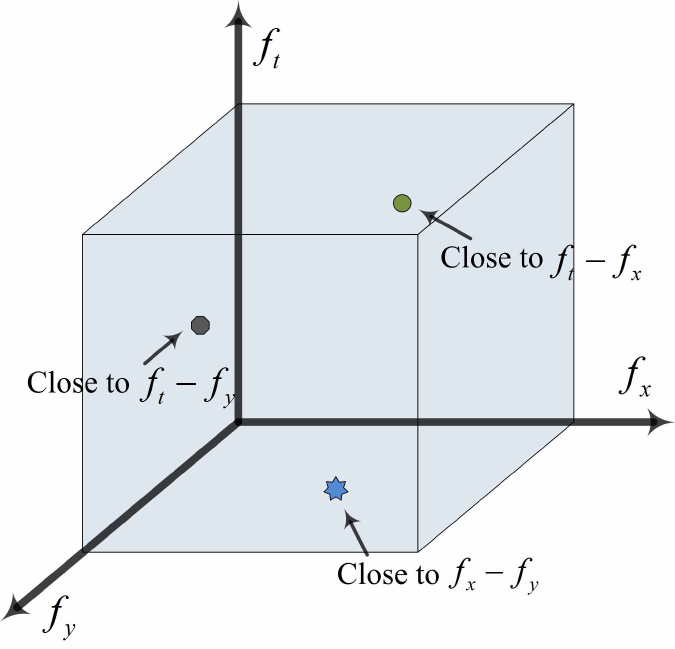}}
  \centerline{(a) Spectral cube.}
\end{minipage}
\hfill
\begin{minipage}[b]{0.49\linewidth}
  \centering
  \hspace{-1cm}
  \centerline{\includegraphics[width=5.2cm]{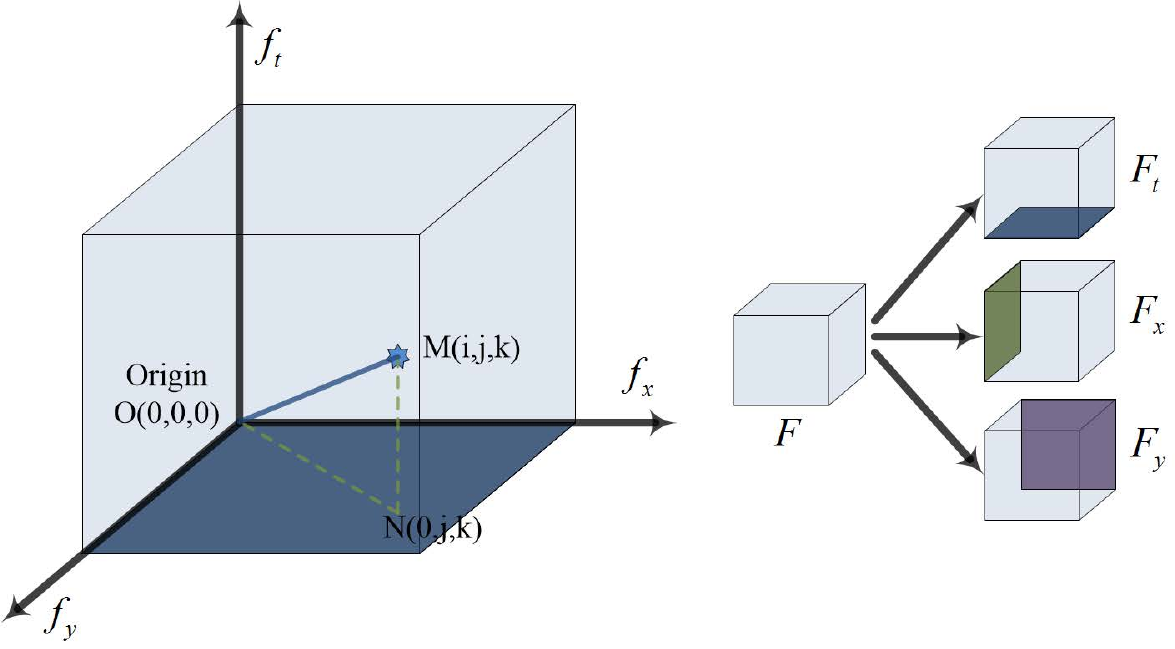}}
  \centerline{(b) Plane projections.} 
\end{minipage}
\caption{The illustration of spectral cube, plane projections, and decompositions.}
\label{fig:Spec_Decomp}
\end{figure}

In the first step, we compute 3D-FFT of $\boldsymbol V$ using a local cube with a sliding window having more than 50\% overlap to yield a volume $\boldsymbol F$. The size of local cube can be adjusted to yield a fine or coarse resolution of the saliency map. In the second step, we perform decompositions of the spectral cube as explained in Fig.~\ref{fig:Spec_Decomp}. Within a 3D spectral cube in $f_t$-$f_x$-$f_y$ coordinate system, if a spectral point is closer to $f_x$-$f_y$-plane, then its projection on $f_x$-$f_y$-plane i.e. along $f_t$-direction will depict variations more prominently as compared to the projections on $f_t$-$f_y$ or $f_t$-$f_x$ planes. Therefore, we decompose the 3D spectral cube by projecting the spectral point $F[i,j,k]$ along $f_t$-direction as
\begin{align}
\label{eqn:Ft}
    \boldsymbol{F_t}[i,j,k] &= \boldsymbol{F}[i,j,k] \times \frac{ON}{OM},\\
    &= \boldsymbol{F}[i,j,k] \times \frac{\sqrt{j^2+k^2}}{\sqrt{i^2+j^2+k^2}},\\
    \boldsymbol{F}[\mu,\nu,\omega]&=\frac{1}{L^3}\sum\limits_{p=0}^{L-1}\sum\limits_{q=0}^{L-1}\sum\limits_{r=0}^{L-1}f[p,q,r]e^{-2\pi i\left(p\mu+q\nu+r\omega \right)/L}, \nonumber
\end{align}
where $[p,q,r]$ and $[\mu,\nu,\omega]$ represent the coordinates in the space and frequency domains, respectively. $L$ defines the size of the local data cube and $f$ is the seismic data within volume $\boldsymbol V$. Similarly, we also compute decompositions along $f_x$- and $f_y$-directions as
\begin{align}
\label{eqn:Fxy}
   & \boldsymbol{F_x}[i,j,k] = \boldsymbol{F}[i,j,k] \times \frac{\sqrt{i^2+k^2}}{\sqrt{i^2+j^2+k^2}},\\
   & \boldsymbol{F_y}[i,j,k] = \boldsymbol{F}[i,j,k] \times \frac{\sqrt{i^2+j^2}}{\sqrt{i^2+j^2+k^2}}.
\end{align}

Thus, after step two, $\boldsymbol F$ is decomposed into three components namely $\boldsymbol F_t$, $\boldsymbol F_x$, and $\boldsymbol F_y$. The equations above for spectral decomposition do not work for a special case, i.e., $i = j = k = 0$. This is the center of 3D spectral cube, which is associated with the DC component of the spectrum and hence does not reflect any changes along three planes. Therefore, we do not include the center point when extracting features from spectral cube.

In the third step, we calculate the absolute mean over each local cube to obtain the corresponding features known as spectral energies, labelled as $\boldsymbol E_t$, $\boldsymbol E_x$, and $\boldsymbol E_y$, respectively. The features extraction process enhance the motion variations along each axis and provide a pixel level descriptions of the energy variations when calculating saliency maps.

\begin{figure}[t]
  \centering
  \includegraphics[width=7.0cm]{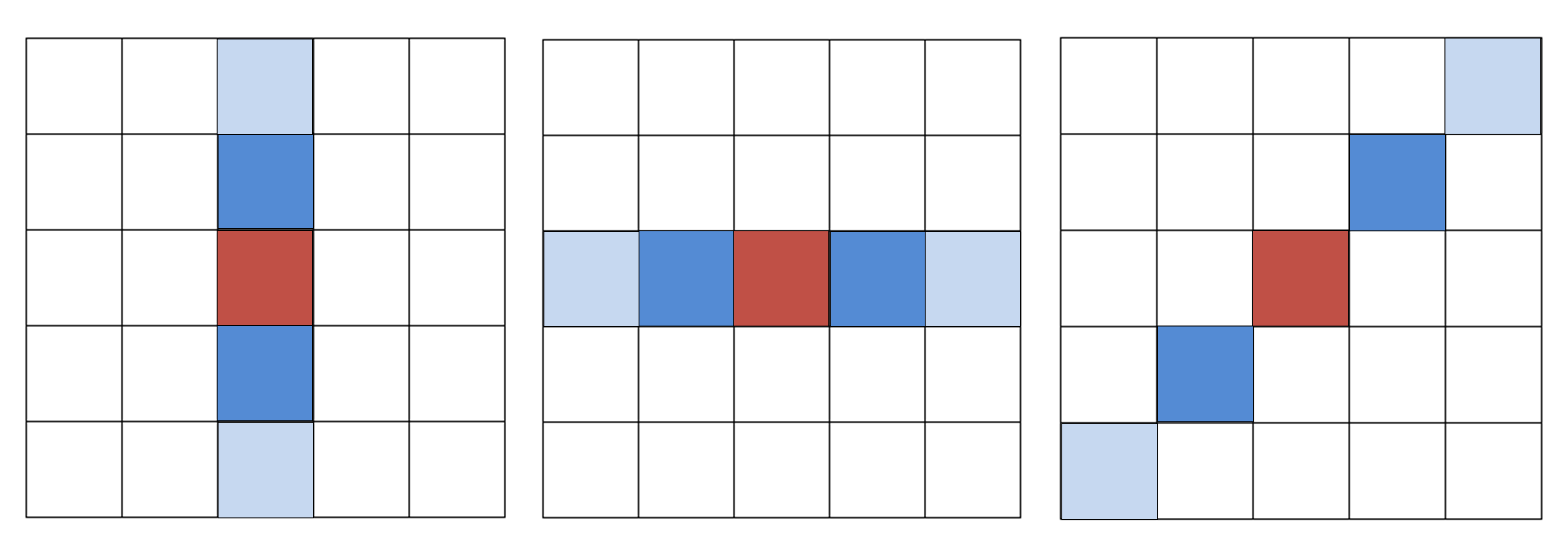}
  \caption{Directional center-surround comparison along $t$, $x$, and $t$-$x$ directions, respectively.}\label{fig:Directional_M}
\end{figure}

The fourth step of the proposed method applies the DCS model to construct the saliency maps $\boldsymbol{S_m}$ using $\boldsymbol{E_m}$ as
\begin{align}\label{eqn:S_x}
\boldsymbol{S_m}[t,x,&y]  = \frac{1}{Q} \sum_{i_0,j_0,r_0} | \boldsymbol{E_m}[t,x,y] - \nonumber \\
& w.\boldsymbol{E_m}[t+i_0,x+j_0,y+r_0] |,~~ m \in \{t,x,y\},
\end{align}
where $i_0$, $j_0$, $r_0$ are chosen such that point $[t+i_0,x+j_0,y+r_0]$ is in the immediate neighborhood of point $[t,x,y]$, such as within a directional window centered at $[t,x,y]$ as depicted in Fig.~\ref{fig:Directional_M}. $Q$ represents the total number of points included in the summation, $w$ represents Gaussian weights, $\boldsymbol{S_m}$ represents $\boldsymbol{S_t}$, $\boldsymbol{S_x}$, or $\boldsymbol{S_y}$, and $\boldsymbol{E_m}$ represents $\boldsymbol{E_t}$, $\boldsymbol{E_x}$, or $\boldsymbol{E_y}$, respectively.

In order to incorporate directionality into saliency calculation and consolidate the top-down salient features, tunable to certain sizes, structures, and orientations, our proposed approach above performs DCS comparison in conjunction with Gaussian weighting of pixel values away from the center point at which it is calculated.~DCS comparisons along $t$, $x$, and $t$-$x$ directions are illustrated in Fig.~\ref{fig:Directional_M}, where dark blue color indicates large Gaussian weights as opposed to small weights for light color pixels. For example, if we are interested in calculating saliency along $x$-direction, then we can not only tune the number of neighboring pixels along $x$-direction (incorporated in DCS computation) but also weights associated with each pixel value. Similarly, we can also tune features directionality (by changing the orientation of DCS comparison) and size (by changing the number of neighboring pixel values included in DCS comparison) in saliency calculation as shown in Fig.~\ref{fig:Directional_M}. In this paper, for DCS comparison, we use a neighborhood window of $d\times1$, where $d$ is the side length of the cube, and apply it along $t$, $x$, and $y$ directions, respectively.

\begin{figure*}[!htb]
  \centering
  \subfigure[A typical seismic inline image.]{\includegraphics[width=.33\textwidth]{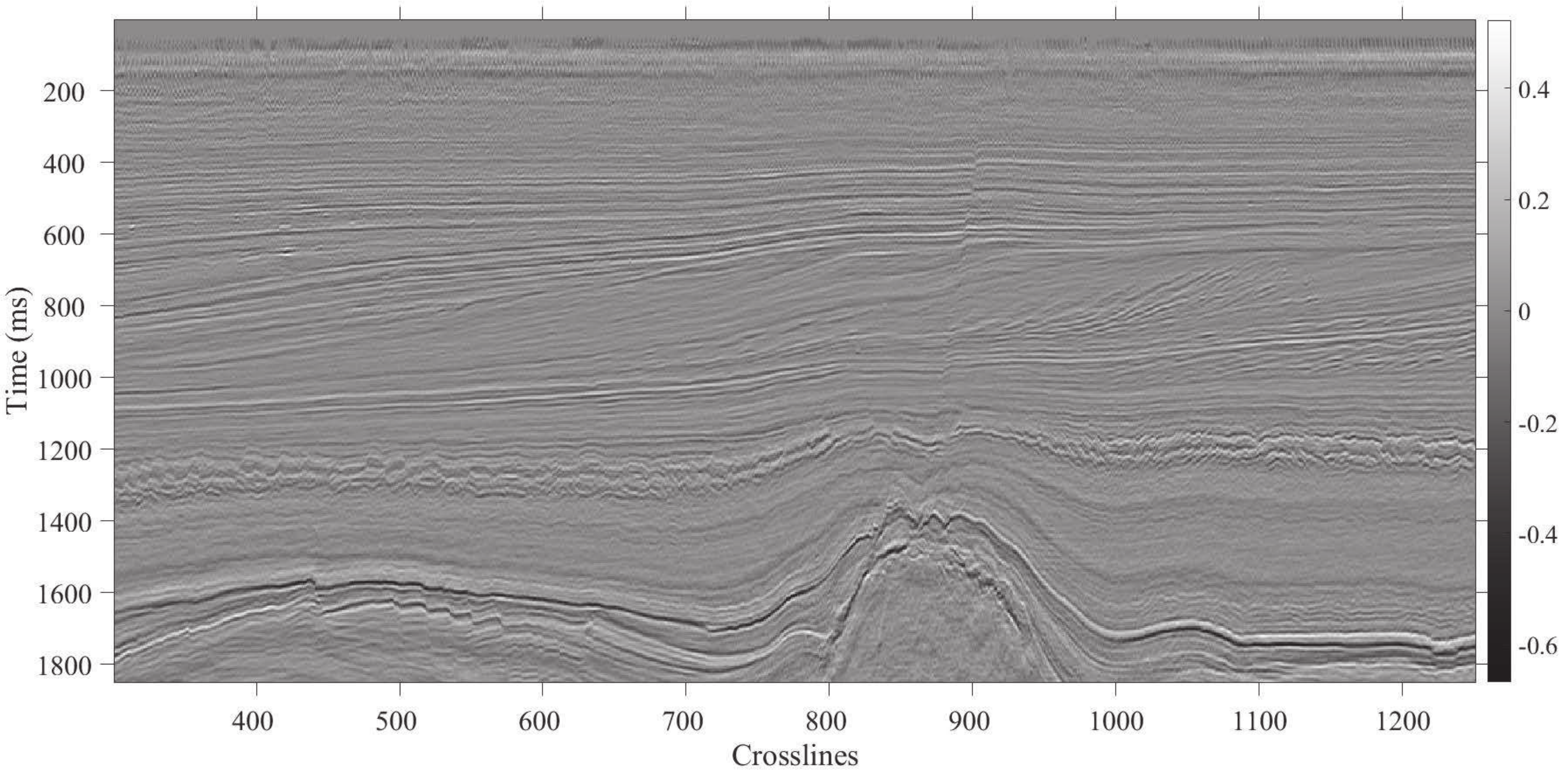}\label{fig:SS245}}
  \subfigure[Zhang et al., (2008) \cite{zhang2008sun}]{\includegraphics[width=.33\textwidth]{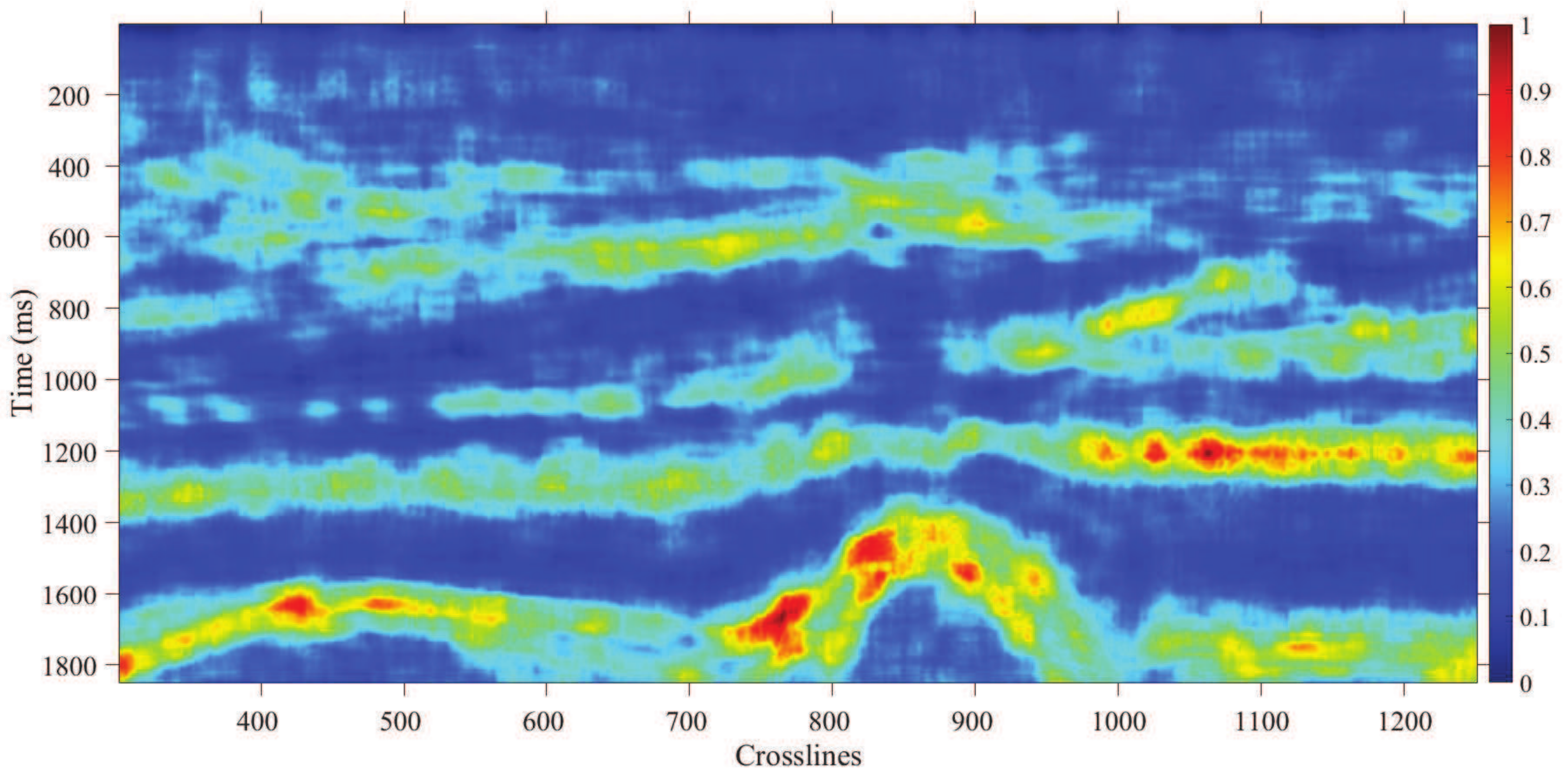}\label{fig:IS}}
  \subfigure[Hou and Zhang, (2007) \cite{hou2007saliency}]{\includegraphics[width=.33\textwidth]{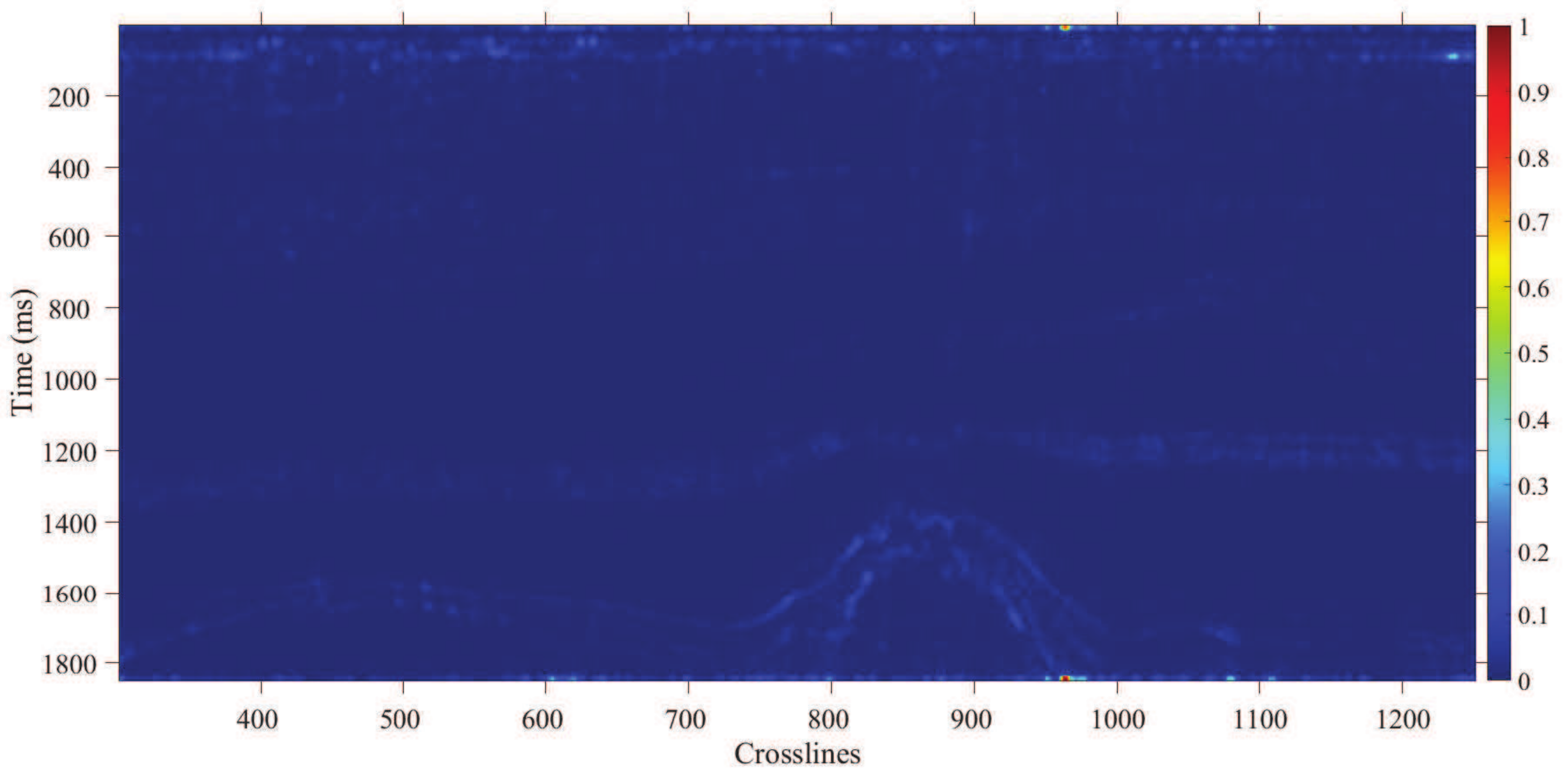}\label{fig:SRes}}
  \subfigure[Guo and Zhang, (2010) \cite{guo2010novel}]{\includegraphics[width=.33\textwidth]{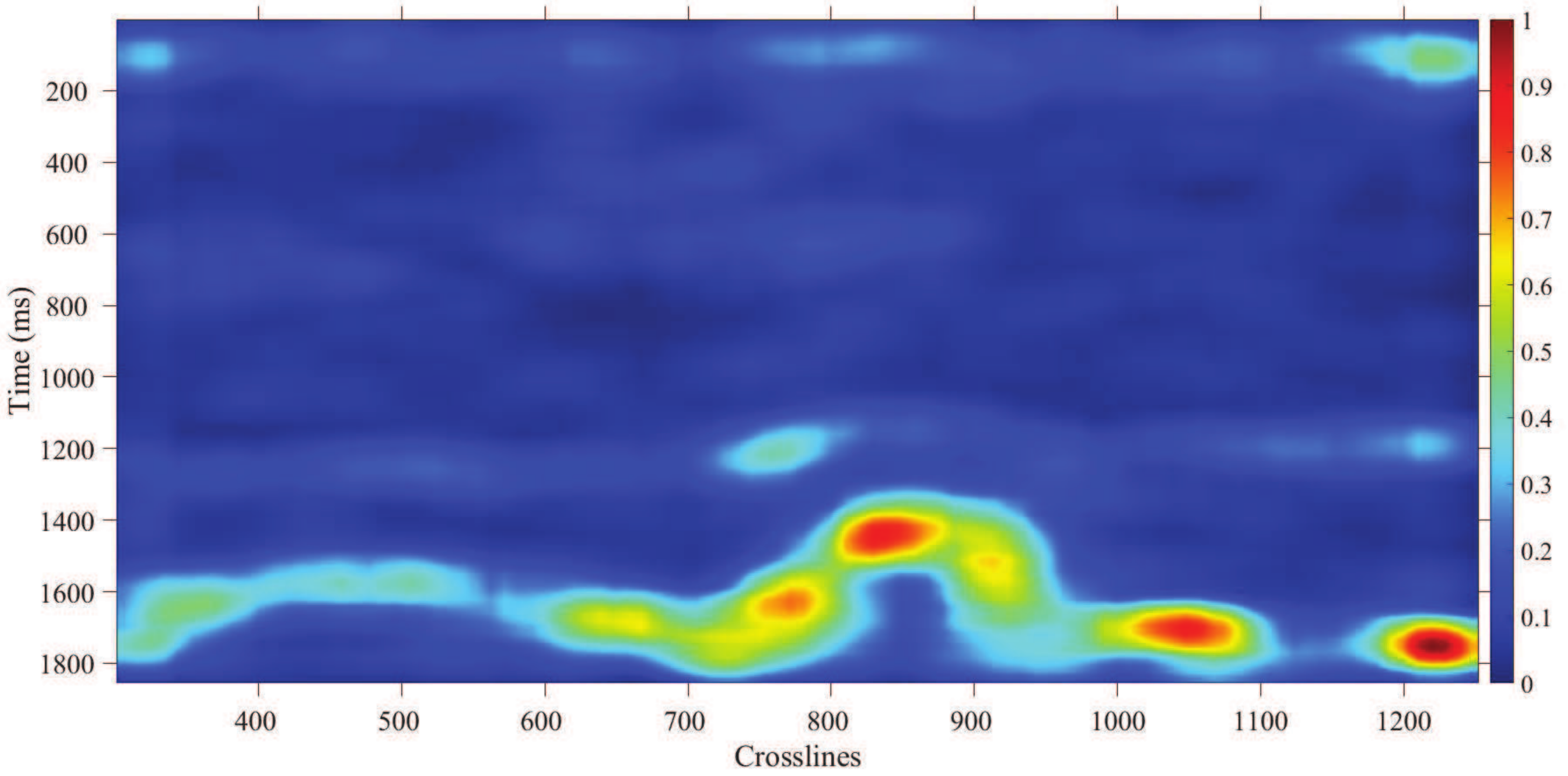}\label{fig:PQFT}}
  \subfigure[Achanta et al., (2008) \cite{achanta2008salient}]{\includegraphics[width=.33\textwidth]{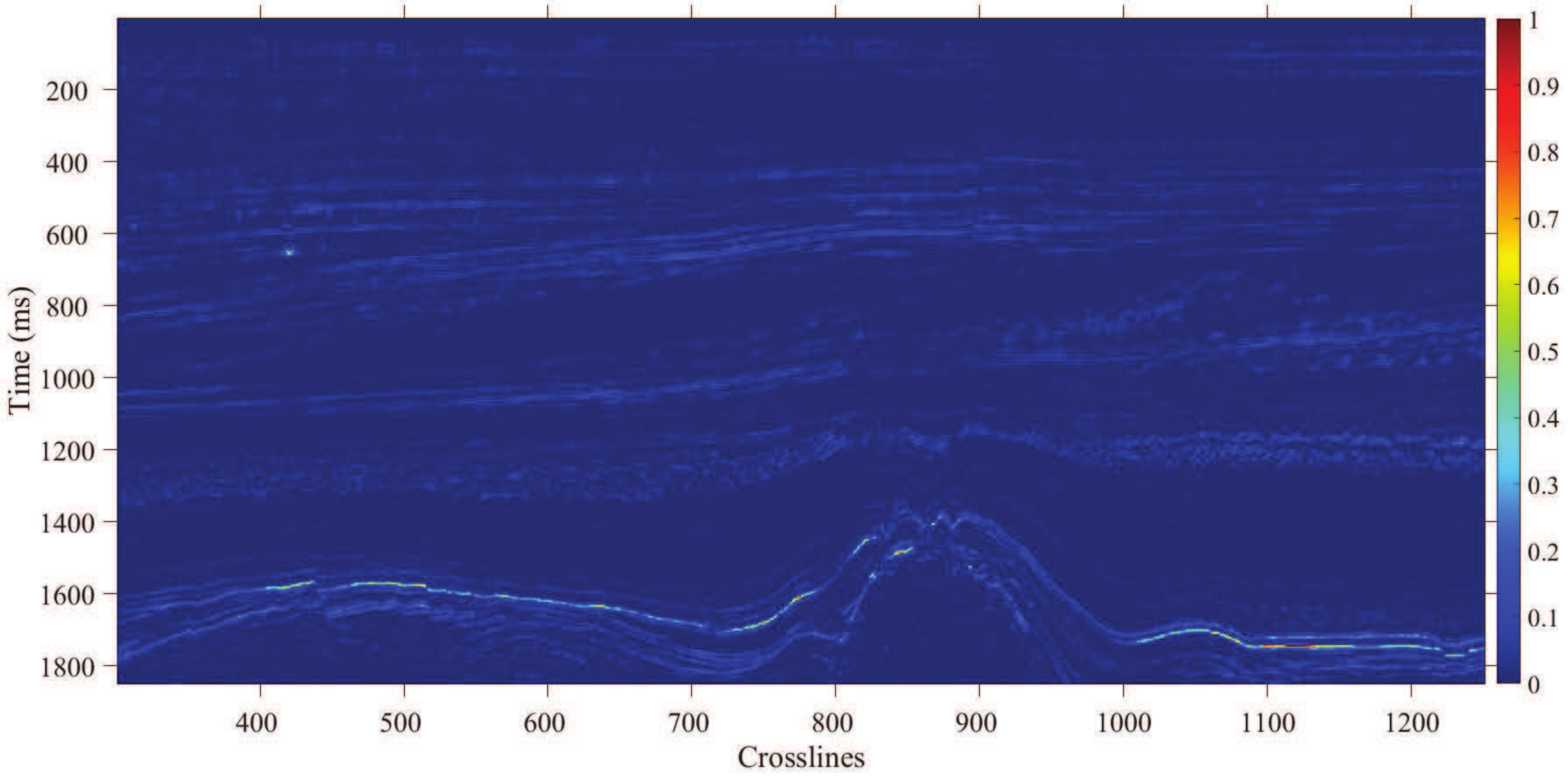}\label{fig:ICVS}}
  \subfigure[Fang et al., (2014) \cite{fang2014video}]{\includegraphics[width=.33\textwidth]{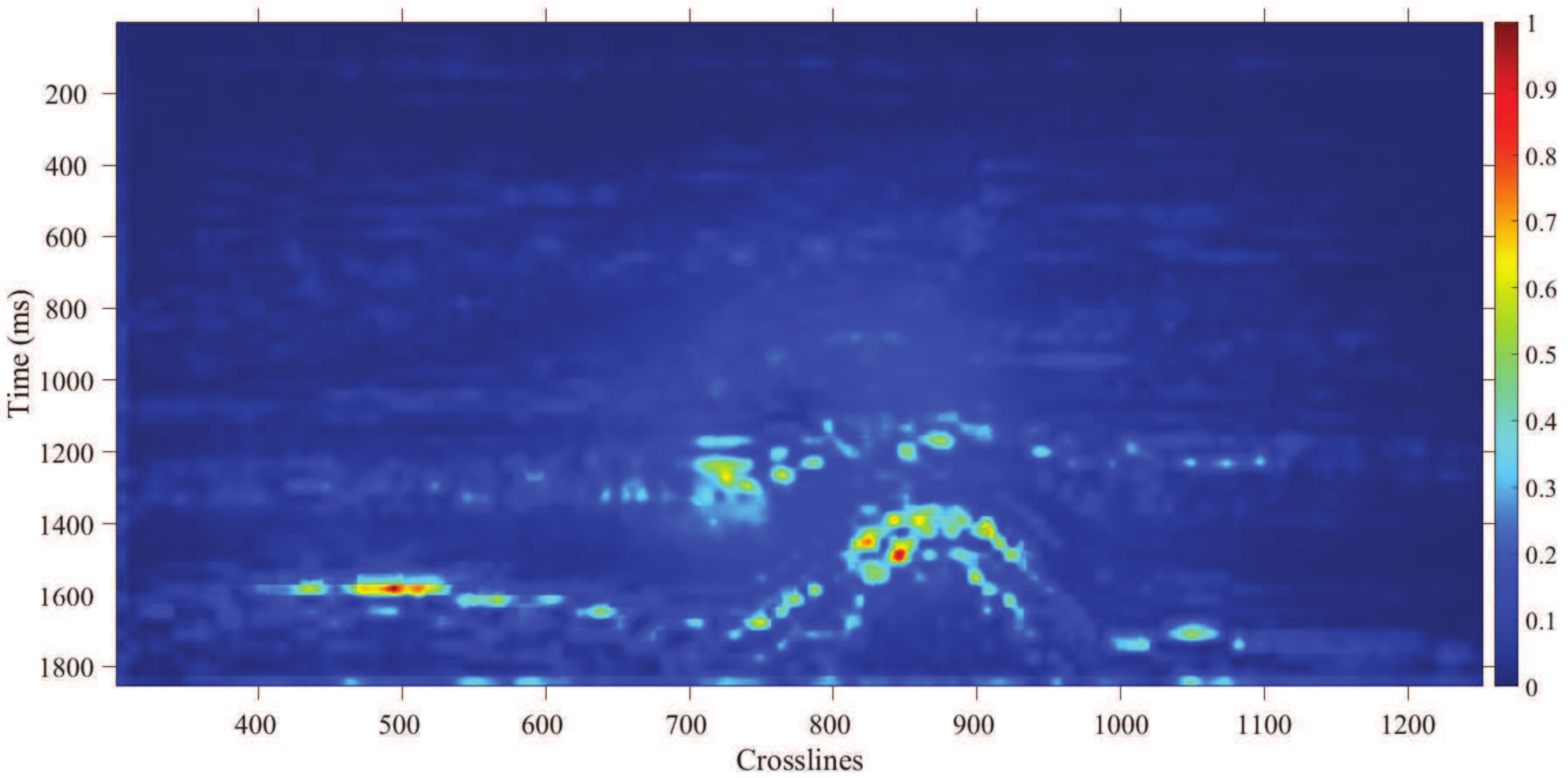}\label{fig:STUW}}
  \subfigure[Seo and Milanfar, (2009) \cite{seo2009static}]{\includegraphics[width=.33\textwidth]{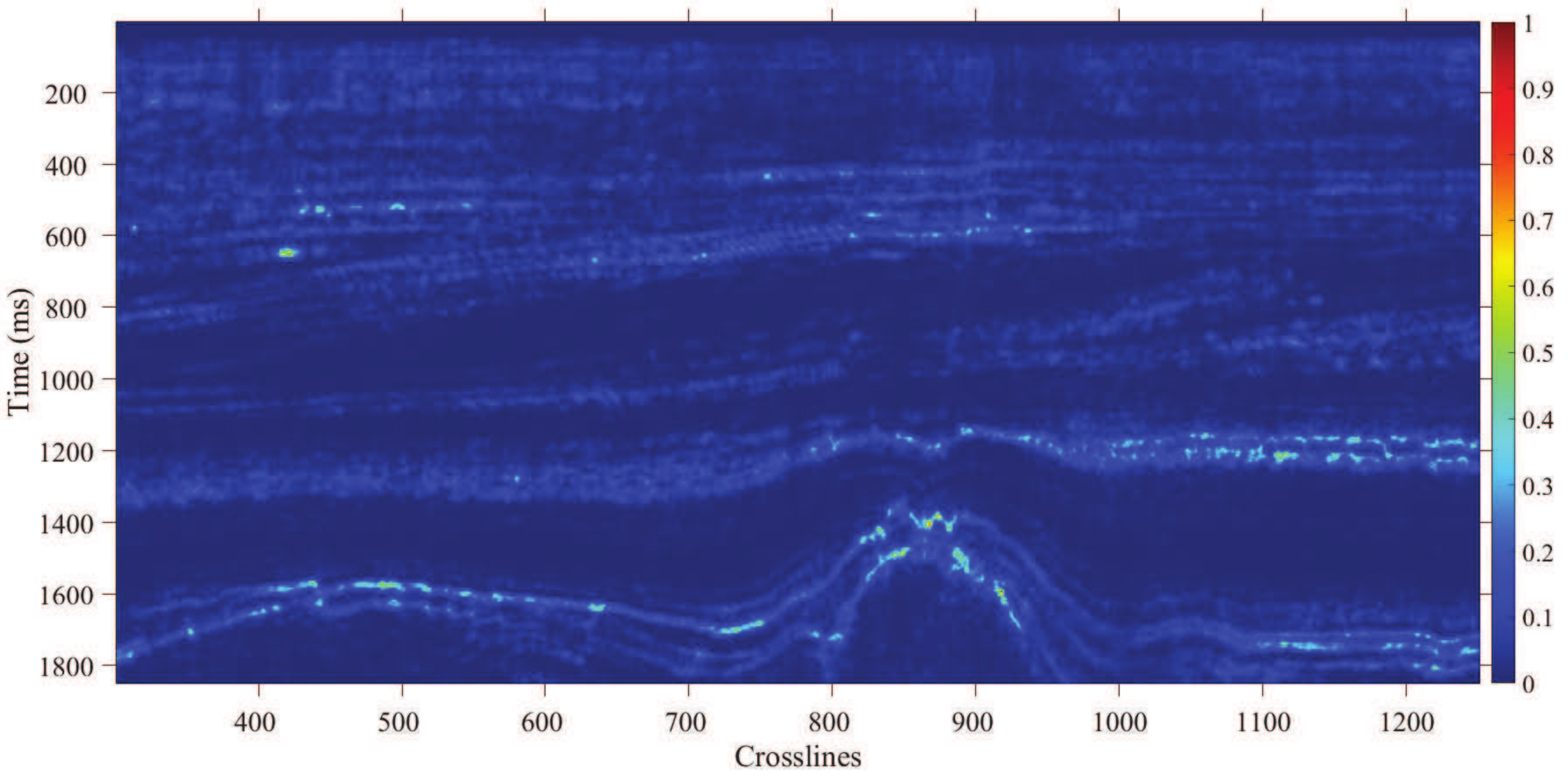}\label{fig:SR}}
  \subfigure[Long and AlRegib, (2015) \cite{Long2015}]{\includegraphics[width=.33\textwidth]{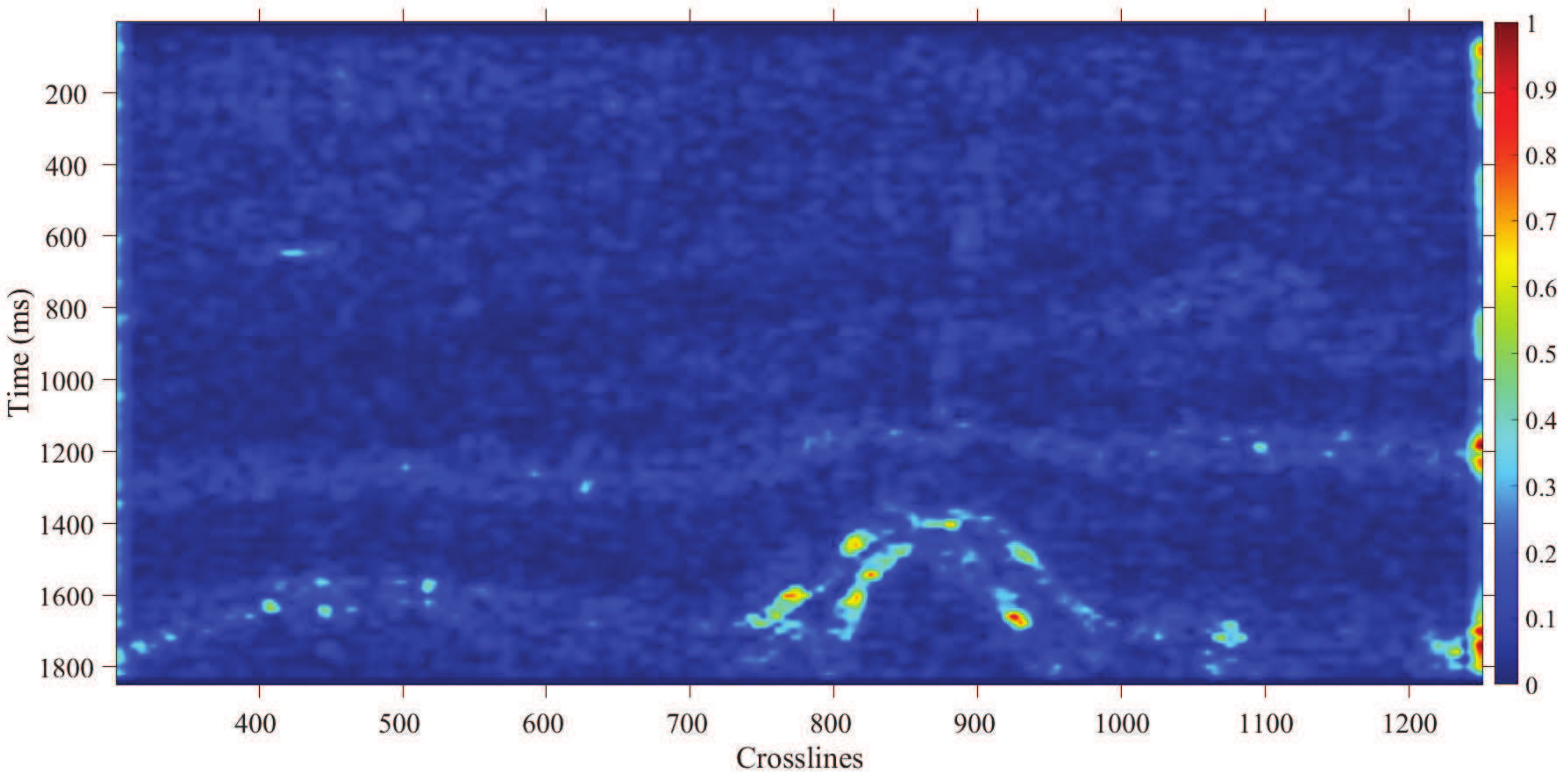}\label{fig:Long}}
  \subfigure[Proposed Method]{\includegraphics[width=.33\textwidth]{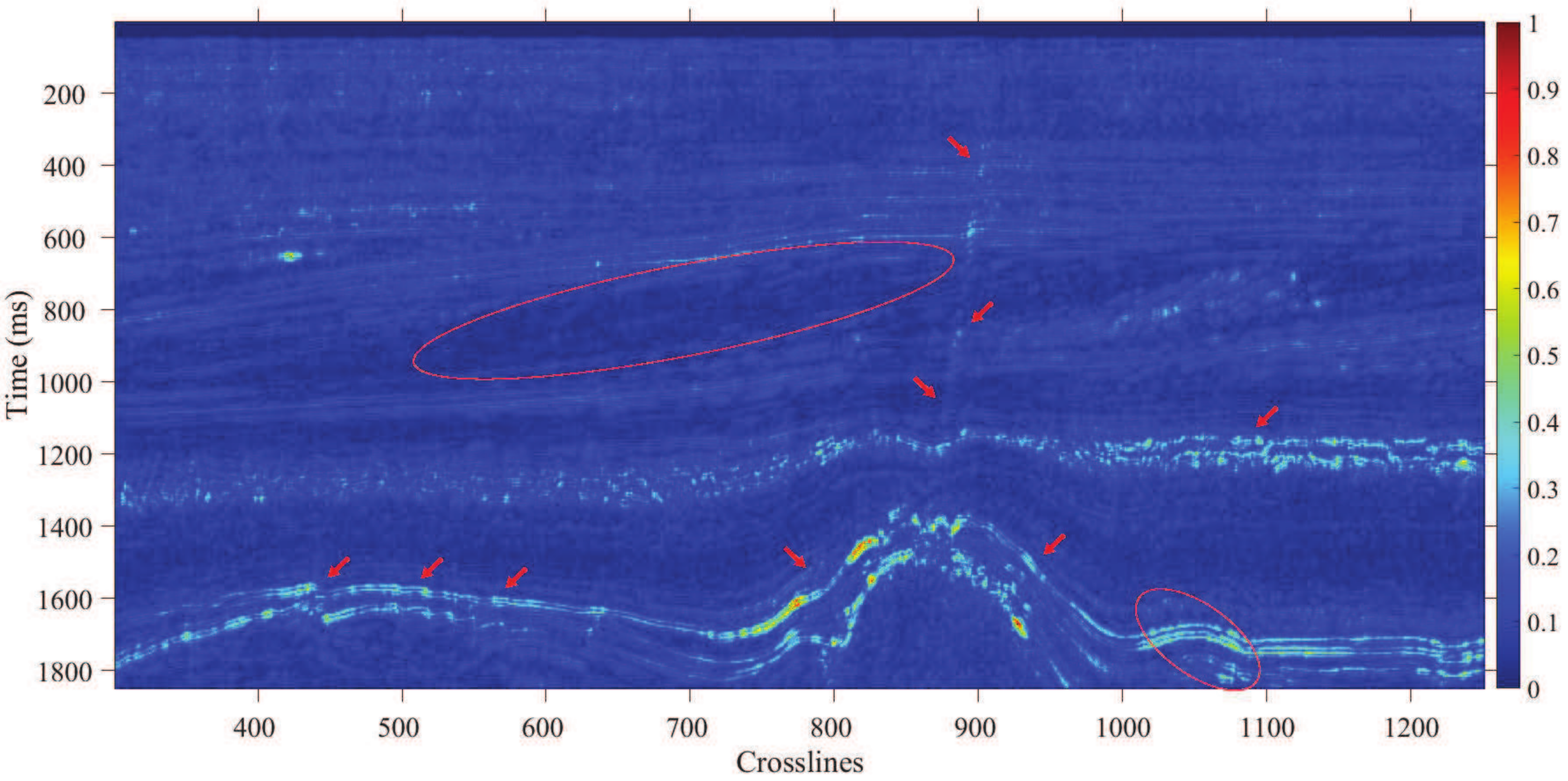}\label{fig:Proposed}}
  \caption{The output of the various saliency detection algorithms on a typical seismic inline section. Red arrows and ellipses highlight the areas, which demonstrate the excellence of the proposed saliency detection algorithm.}
  \label{fig:SaliencyMaps}
\end{figure*}

Finally, the saliency map $\boldsymbol{S}$, which is of the same size as that of $\boldsymbol{V}$ is obtained as
\begin{equation} \label{eqn:S}
\boldsymbol{S}[i,j,k]= a \cdot \boldsymbol{S_t}[i,j,k] + b \cdot \boldsymbol{S_x}[i,j,k] + c \cdot \boldsymbol{S_y}[i,j,k].
\end{equation}
The weights in saliency calculation i.e. $a$, $b$, and $c$ can be set either equally to construct a saliency map with equal distribution of saliency maps calculated along $t$, $x$, and $y$ directions or empirically to highlight certain features along any particular direction. In this work, we have used a cube of size $5\times5\times5$ and equal weights $a$, $b$, and $c$ for saliency calculation. The proposed saliency detection is based on 3D-FFT, which makes it fast and obtains saliency maps with adjustable resolution by varying the cube size. Furthermore, the proposed approach is computationally inexpensive and requires very few parameters as compared to other visual saliency algorithms.

\vspace{-.2cm}
\section{Results}
\label{sec:Results}
\vspace{-.2cm}
In this section, we present the results of saliency detection on a real seismic dataset acquired from the Netherlands offshore F3 block in the North Sea whose size is $24\times 16\mbox{ km}^2$. A typical seismic inline section from this dataset containing multiple seismic facies is shown in Fig.~\ref{fig:SaliencyMaps}a. A well-founded saliency algorithm can not only resolve spatial variations along different directions within seismic volume but also highlight the contrast of different geological structures with respect to its surrounding sediments. The results of the state-of-the-art image and video saliency detection algorithms presented in \cite{zhang2008sun}, \cite{hou2007saliency}, \cite{guo2010novel}, \cite{achanta2008salient}, \cite{fang2014video}, \cite{seo2009static}, \cite{Long2015}, and the proposed method are shown in Fig.~\ref{fig:SaliencyMaps}b-i, respectively.

Subjective evaluation of the results show that the proposed method effectively highlights salient features from a seismic image as compared to other state-of-the-art algorithms. Specifically, red arrows and ellipses in Fig.~\ref{fig:SaliencyMaps}i highlight regions in a seismic inline section, which demonstrate the excellence of proposed saliency detection as opposed to other saliency detection algorithms. As observed in Fig.~\ref{fig:SaliencyMaps}, a majority of algorithms fail to detect a major fault in the center of a seismic image. Such kind of faults are characterized by subtle variations in intensity and texture, which make them extremely challenging and difficult to highlight using a small set of seismic attributes. Figure~\ref{fig:SaliencyMaps}i shows that such fault is adequately highlighted by the proposed saliency detection method because it takes into account the spectral variations along all three dimensions of the seismic data. In addition, the proposed saliency detection algorithm pleasantly suppresses a sigmoidal structure with respect to its surrounding, indicated by an ellipse in the middle of seismic section, which is not distinctively detected by other algorithms. Furthermore, the amplitude of salient values detected by the proposed algorithm near the salt-dome boundary are not only more localized but also significantly higher than most of other state-of-the-art algorithms. Similarly, red arrows in the bottom left and middle right portion of the proposed saliency map highlight areas such as smaller faults and chaotic structures that are not clearly visible in other saliency maps. Finally, it can be observed from Fig.~\ref{fig:SaliencyMaps} that the resolution of the proposed approach is much better as compared to that of other saliency detection algorithm, which makes it advantageous for applications such as seismic interpretation, which requires not only fine perception but also efficient detection of subtle features from images and videos. Therefore, the proposed approach is expected to not only become a very handy tool for interpreter-assisted seismic interpretation but can also serve as a base attribute map for creating workflows for automated detection of various geological structures.

\vspace{-.2cm}
\section{Conclusion}
\label{sec:conclusion}
\vspace{-.2cm}
In this paper, we have developed a new saliency detection algorithm for seismic applications using features based on 3D-FFT local spectra and multi-dimensional plane projections. We have proposed a novel approach for feature extraction based on spectral cube coupled with directional center-surround model to estimate the salient features effectively. The proposed algorithm is based on 3D-FFT, which makes it advantageous for application on large datasets, and computationally inexpensive and real-time implementations. Simulation results on a real seismic dataset show the efficacy of the proposed scheme in the detection of salient points and subtle features observed in a geologically complex setting. Furthermore, experimental results also show that the proposed method outperforms the state-of-the-arts methods for saliency detection in seismic applications.

\bibliographystyle{IEEEbib}
\bibliography{main}

\end{document}